\documentclass[numsec,webpdf,modern,medium,namedate]{article}
\onecolumn
\usepackage{booktabs}  
\usepackage{caption}
\usepackage{array}
\usepackage{amsfonts}
\usepackage{amsmath}          
\usepackage{amsthm}           
\usepackage{tikz}
\usepackage{graphicx}         
\usepackage{enumerate}        
\usepackage{natbib}           
\usepackage{url} 

\usepackage{algorithm}
\usepackage{algorithmicx}
\graphicspath{{Fig/}}
\usepackage{algpseudocode}
\theoremstyle{plain} 
\newtheorem{theorem}{Theorem} 
\newtheorem{proposition}[theorem]{Proposition}
\newtheorem{corollary}[theorem]{Corollary} 
\newtheorem{assumption}[theorem]{Assumption} 

\theoremstyle{definition}
\newtheorem{definition}{Definition}

\theoremstyle{remark}

\newtheorem{remark}{Remark}

\usepackage{authblk}

\begin{document}

\title{Model-Free Kernel Conformal Depth Measures Algorithm for Uncertainty Quantification in Regression Models in Separable Hilbert Spaces}

\author[1]{Marcos Matabuena\thanks{*Corresponding author}}
\author[2]{Rahul Ghosal}
\author[3]{Pavlo Mozharovskyi}
\author[4]{Oscar Hernan Madrid Padilla}
\author[1]{Jukka-Pekka Onnela}

\affil[1]{Department of Biostatistics, Harvard University\\Boston, MA 02115, USA}
\affil[2]{Department of Epidemiology and Biostatistics, University of South Carolina\\Columbia, SC 29208, USA}
\affil[3]{LTCI, Télécom Paris, Institut Polytechnique de Paris\\Palaiseau 91120, France}
\affil[4]{Department of Statistics, UCLA\\Los Angeles, CA 90095, USA}


\maketitle

\begin{abstract}
    
Depth measures are powerful tools for defining level sets in emerging, non-standard, and complex random objects such as high-dimensional multivariate data, functional data, and random graphs. Despite their favorable theoretical properties, the integration of depth measures into regression modeling to provide prediction regions remains a largely underexplored area of research.
To address this gap, we propose a novel, model-free uncertainty quantification algorithm based on conditional depth measures—specifically, conditional kernel mean embeddings and an integrated depth measure. These new algorithms can be used to define prediction and tolerance regions when predictors and responses are defined in separable Hilbert spaces. The use of kernel mean embeddings ensures faster convergence rates in prediction region estimation.
To enhance the practical utility of the algorithms with finite samples, we also introduce a conformal prediction variant that provides marginal, non-asymptotic guarantees for the derived prediction regions. Additionally, we establish both conditional and unconditional consistency results, as well as fast convergence rates in certain homoscedastic settings.
We evaluate the finite-sample performance of our model in extensive simulation studies involving various types of functional data and traditional Euclidean scenarios. Finally, we demonstrate the practical relevance of our approach through a digital health application related to physical activity, aiming to provide personalized recommendations. 
\end{abstract}

\section{Introduction}
In recent years, the study and application of depth measures \cite{zuo2000general} for various statistical modeling tasks, such as exploratory analysis \cite{liu1999multivariate} and classification \cite{li2012dd}, have attracted significant attention in the statistical community (see, e.g., \cite{LopezPintadoR09, NarisettyN16}; see also \cite{MoslerM22} for a recent survey). From a practical standpoint, depth measures serve as powerful alternatives to traditional centrality concepts such as the mean and median. Additionally, depth measures offer advantageous mathematical properties, such as robustness to outliers and invariance under affine transformations, that facilitate the definition of order statistics for multivariate data and other high-dimensional random objects, including mathematical functions. This is particularly valuable in contexts where a natural canonical notion of order is absent, such as with bivariate data or infinite-dimensional functional spaces.

The concept of depth was first introduced by Mahalanobis and Tukey (see, for example, \cite{mclachlan1999mahalanobis, berrendero2020mahalanobis, tukey1975mathematics}), with an important historical background in bivariate hypothesis testing rank tests \cite{hodges1955bivariate}. Subsequent theoretical developments have explored the robustness properties of depth functions, particularly in relation to breakdown points and influence functions \cite{donoho1982breakdown, donoho1992breakdown}.

The primary purpose of statistical modeling with depth measures is to define a rank for an element 
\( y \in \mathcal{Y} \), denoted by \(\operatorname{Rank}(y,\mathcal{D}_{n})\), where random sample
\(\mathcal{D}_{n} = (Y_{1}, \dots, Y_{n})\) is an independent and identically distributed (i.i.d.) 
sequence of random elements drawn from a distribution \( P \). This rank measures the centrality of a new data point 
\( y \) relative to the reference distribution \( P \). Values of \(\operatorname{Rank}(y,\mathcal{D}_{n})\) 
close to one indicate points near the geometric center of the distribution, while values close to zero 
correspond to more extreme points.

Early depth measures ranked points according to their Euclidean distance from the geometric center of the data distribution—analogous to the multivariate mean or median—as exemplified by the Mahalanobis distance. However, this approach has limitations when applied to non-Gaussian distributions, such as those that are multimodal or asymmetric, where a simple distance fails to capture the topological and geometrical differences between two distributions. To address these challenges, a second wave of depth research emerged in the 1980s and 1990s, introducing new generations of depth measures such as Oja depth, simplicial depth, and majority depth. These new depths are able to detect a richer variety of distributional shapes, such as multimodal distributions. For a comprehensive review, see \cite{liu1990notion, oja1983descriptive, singh1991notion, liu1999multivariate, zuo2000general}.

Depth measures gained prominence in the 1970s for exploratory analysis purposes, with their application now extending to outlier detection \cite{chen2008outlier}, variable selection, hypothesis testing \cite{singh2022some, liu1999multivariate}, clustering , missing data, and classification \cite{li2012dd}. The robustness of depth measures, particularly their role in robust statistics \cite{denecke2012consistency} in terms of influence functions \cite{dang2009influence}, has been a significant area of theoretical analysis in recent years.

Currently, a third wave of research on depth measures is underway, motivated by technological improvements in data collection. This wave aims to define general depth measures applicable to a variety of statistical random objects, including functional data, high-dimensional data, and non-Euclidean data types such as random graphs and multivariate density functions residing in nonlinear metric spaces \cite{virta2023spatial, geenens2023statistical, dai2023tukey, LafayeDeMicheauxMV20}. These new methodologies for analyzing random objects in abstract spaces are particularly relevant in modern healthcare and public health applications \cite{rodriguez2022contributions, lugosi2024uncertainty}, facilitating deep patient phenotyping and the analysis of complex data structures such as biosensor signals and medical images. In diabetes research \cite{matabuena2021glucodensities}, for instance, continuous glucose monitoring data may be used to identify anomalous low and high glucose events—clinically corresponding to hypoglycemia and hyperglycemia events.

In a seminal work, Li et al. \cite{li2008multivariate} propose novel methodologies to define level sets and tolerance regions based on the mathematical notion of depth measures, albeit from an unconditional perspective. Tolerance regions—statistical regions expected to contain a specified proportion of individuals in a population with a certain probability—are valuable tools in data analysis across various fields. Particularly in quality control and medical science, they are used to establish normal ranges for clinical biomarkers derived from healthy populations \cite{murphy1948non, krishnamoorthy2009statistical}. These regions facilitate the classification of patients into distinct categories, such as healthy and diseased groups, by providing threshold values that distinguish between normal and abnormal biomarker levels \cite{young2020nonparametric}.

Depth measures have been shown to outperform traditional nonparametric kernel density estimation methods when reconstructing level sets of data distributions under various conditions and data structures, including functional spaces \cite{tsybakov1997nonparametric, LopezPintadoR09}. In particular, depth measures—through the use of depth bands—demonstrate robustness against the curse of dimensionality and outliers, and they achieve fast convergence rates in many scenarios (see, for instance, \cite{pmlr-v70-jiang17b} and \cite{zhang2002some}).

Conformal prediction \citep{shafer2008tutorial} is an increasingly popular analytic framework to quantify uncertainty in predictive modeling. Given an independently and identically distributed (i.i.d.) random sample \( \mathcal{D}_n = \{(X_i, Y_i) \}_{i=1}^n  \subset \mathcal{X} \times \mathcal{Y} \) and a arbitrary regression model \( m: \mathcal{X} \to \mathcal{Y} \), conformal inference provides a general framework to construct an estimated prediction region \( \widehat{C}^\alpha(\cdot) \) that offers non-asymptotic guarantees  \citep{shafer2008tutorial, barber2023conformal} of the type:
\[
\mathbb{P}(Y \in \widehat{C}^\alpha(X)) \geq 1 - \alpha,
\]
\noindent where \( \mathbb{P} \) is the probability law over the random sample \( \mathcal{D}_n \) and the random pair \( (X, Y) \). However, the development of conformal prediction for settings with non-standard data structures—such as functional data and graphs—remains limited \citep{lugosi2024uncertainty}. This limitation stems primarily from the difficulties involved in efficiently defining ranking methods in general spaces. One potential approach to overcoming this challenge is to incorporate depth measures into conformal prediction algorithms.

To address this gap, we propose a novel framework that integrates depth measures to define rankings in general spaces, and then combines them with conformal inference to produce powerful prediction regions equipped with non-asymptotic guarantees. From an asymptotic standpoint, our methods achieve fast nonparametric convergence rates by incorporating kernel mean embeddings as depth measures \citep{muandet2017kernel}. From a literature and technical perspective, our approach extends to multivariate and random objects defined in separable Hilbert space models within the distributional conformal framework introduced by \cite{chernozhukov2021exact} for univariate outcomes.

\subsection{Motivating Application: Personalization of Physical Activity}

The digital and precision medicine revolution has sparked a transformative era of artificial intelligence in healthcare, bringing forth advanced and personalized clinical interventions \cite{kosorok2019precision, matabuena2024multilevel}. This paradigm shift empowers healthcare providers to harness modern wearable devices for continuous monitoring of patient characteristics, opening avenues for more sophisticated phenotyping and tailored treatments. Physical activity is a domain that stands to greatly benefit from these technological advancements \cite{matabuena2022physical}.

From a clinical perspective, physical activity represents a low-cost intervention \cite{bolin2018physical, mendes2016exercise, franks2017causal, friedenreich2021physical, burtscher2020run} with significant potential to positively impact human health across various diseases. The integration of wearable devices enables comprehensive, high-resolution tracking of patients' physical activity, allowing healthcare professionals to gain deeper insights into patients' overall health and wellness, as well as to personalize physical activity interventions.

\begin{figure}[ht!]
  \centering
  \includegraphics[width=0.45\textwidth]{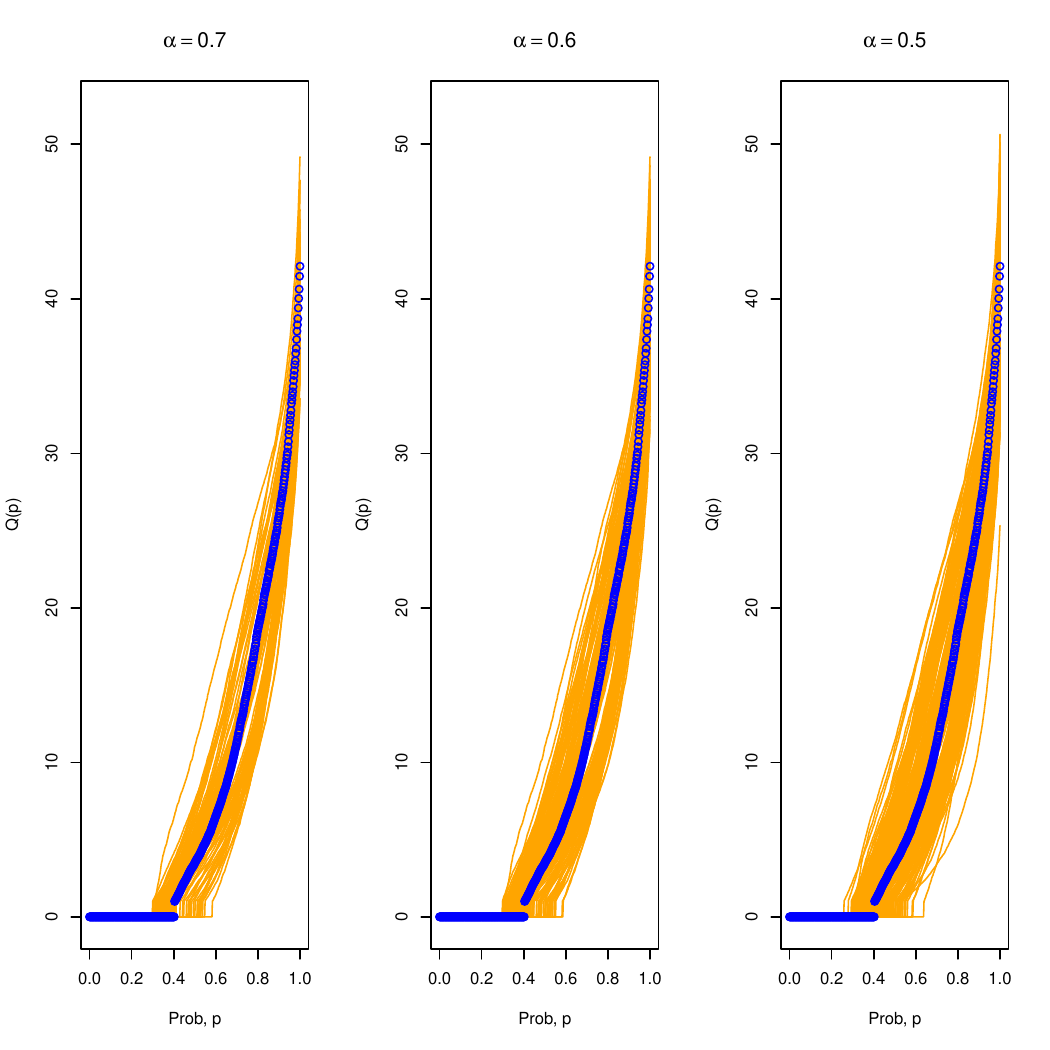}%
  \hspace{1em} 
  \includegraphics[width=0.45\textwidth]{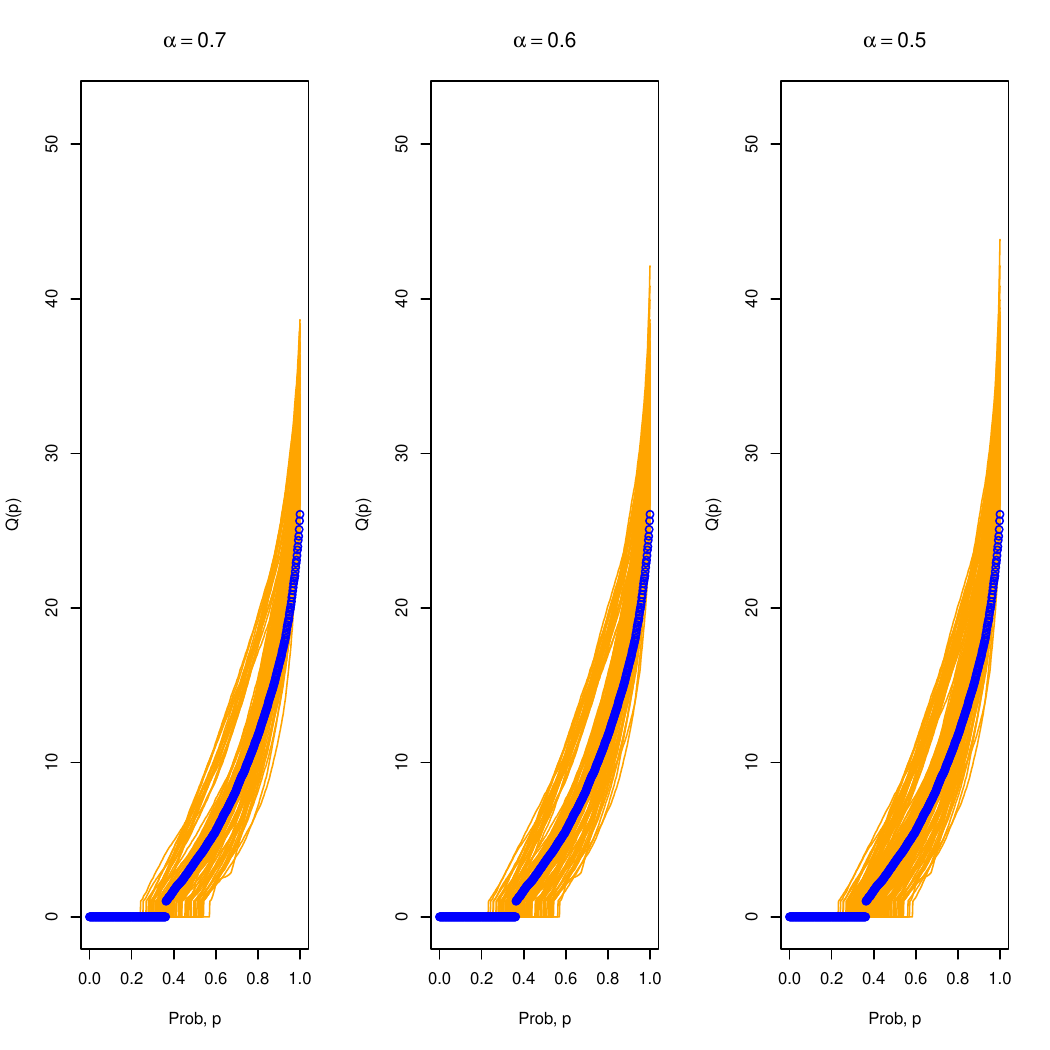}
  \caption{Physical activity predictive regions for the individuals a) and b)}
    \label{fig:papera}

\end{figure}

This study is motivated by the need for more personalized physical exercise recommendations based on individual characteristics, utilizing a novel prediction regions algorithm. Although there is a growing movement towards personalized physical activity prescriptions, current medical guidelines suggests popululation bases--standardized recommendations, such as weekly aerobic exercise ranging from 150 to 300 minutes \cite{piercy2018physical}. To illustrate the intuition behind our method, Figure \ref{fig:papera} displays the observed quantile trajectories (in blue) of physical activity for two female individuals, along with the estimated prediction regions at various confidence levels \( \alpha \in (0,1) \). These regions are calculated using healthy individuals from the U.S. population as a reference. We observe that individual (a) and (b) falls within the expected range. By creating prediction regions conditioned on the characteristics of the patients, we provide a formal method to actualize individualized recommendations of physical activity.
    

    
    
    
    

\subsection{Notation and Problem Definition}

Consider a random sample $\mathcal{D}_{n}=\{(X_1,Y_1),\dots,(X_n,Y_n)\}$ of $n$ independent copies of the random variables $(X,Y)\sim P$, where $X$ takes values in the space $\mathcal{X}$ and $Y$ takes values in the space $\mathcal{Y}$. Here, we assume that $\mathcal{X}$ and $\mathcal{Y}$ are separable Hilbert spaces. Let $k_{\mathcal{X}}: \mathcal{X} \times \mathcal{X} \rightarrow \mathbb{R}$ be a positive definite kernel on $\mathcal{X}$, inducing a reproducing kernel Hilbert space (RKHS) $\mathcal{H}_{\mathcal{X}}$ with an inner product $\langle \cdot, \cdot \rangle_{\mathcal{X}}$. Similarly, let $k_{\mathcal{Y}}: \mathcal{Y} \times \mathcal{Y} \rightarrow \mathbb{R}$ be defined, and denote the related RKHS as $\mathcal{H}_{\mathcal{Y}}$. To avoid technical problems, we assume that $k_{\mathcal{X}}$ and $k_{\mathcal{Y}}$ are universal kernels on $\mathcal{X}$ and $\mathcal{Y}$, respectively, and then, $\mathcal{H}_{\mathcal{X}}$, $\mathcal{H}_{\mathcal{Y}}$ are simultaneously dense on $\mathcal{X}$ and $\mathcal{Y}$, see \cite{sriperumbudur2011universality}.

In this paper, for simplicity, we restrict our attention to a regression model defined as follows:

\begin{equation}\label{eqn:model}
    Y= m(X)+\sigma(X)\epsilon= m(X)+\epsilon(X),
\end{equation}

\noindent where $\epsilon(\cdot) \in \text{GP}(0_{\mathcal{H}_{\mathcal{Y}}},\Sigma)$ is the random error specified by a Gaussian process, $m(\cdot)\in  \mathcal{H}_{\mathcal{Y}}$ is the regression function, and $\sigma(X)\in \mathcal{H}_{\mathcal{Y}}$  introduces the heteroscedastic nature in the random error. In the case where $\sigma(X)= \sigma\in \mathbb{R}^{+},$ we say that the regression model defined in Equation \ref{eqn:model} is homoscedastic, and we denote $\epsilon(X)$ simply as $\epsilon$. For identification purposes, we suppose that $\mathbb{E}(\epsilon(X)|X)=0$ and $\text{Cov}(\epsilon(X),\epsilon(X))=\Sigma(X)$.

In our specific context, we define the region of probability level population prediction $\alpha \in (0,1)$ as a subset of the joint space $\mathcal{X} \times \mathcal{Y}$ that holds the true response $Y$ for a given predictor $X$ with probability $1-\alpha$, that is, $P(Y\in C^{\alpha}(X)|X)=  1 - \alpha$ for all $(X,Y) \in \mathcal{X} \times \mathcal{Y}$. For simplicity, given a random pair $(X,Y)$, we assume that such a predictive region exists. For practical purposes, we establish the convergence to certain oracle prediction regions that we also denote by $C^{\alpha}(X)$ (that we define later).



Our proposed algorithms employ data‐splitting to improve calibration, so we introduce some additional notation. In order to estimate the predictive region \(C^{\alpha}(x)\),  we partition the observed random sample $\mathcal{D}_{n}$ into three disjoint subsets: $\mathcal{D}_{n}=\mathcal{D}_{\text{train}} \cup \mathcal{D}_{\text{train}{2}} \cup \mathcal{D}_{\text{test}}$. The index sets are denoted as $\mathcal{S}_{1}:= \{ i\in [n]: (X_i, Y_i)\in \mathcal{D}_{\text{train}} \}$, $\mathcal{S}_{2}:= \{ i\in [n]: (X_i, Y_i)\in \mathcal{D}_{\text{train}{2}} \}$, $\mathcal{S}_{3}:= \{ i\in [n]: (X_i, Y_i)\in \mathcal{D}_{\text{test}} \}$, where $[n]:= \{1,\dots, n\}$.

\subsection{Summary of Contributions}
The main contributions of the present work are discussed bellow:
\begin{enumerate}
    \item  We present a novel methodology for uncertainty quantification, based on the notion of conditional kernel mean embeddings \cite{muandet2017kernel}. 
    Our algorithms operate when the response and predictors take values in a separable Hilbert space, or in more general metric spaces of negative type \cite{schoenberg1937certain,schoenberg1938metric}. To illustrate the operational capacities of our methods for random objects in abstract spaces, we focus on a clinical application that considers as an outcome the probability distribution of physical activity time series. Here, we employ the 2-Wasserstein distance as a similarity measure between the probability distributions, which is equivalent to using the quantile functions derived from physical activity probability distributions.
    
    \item  We propose conformalized versions \citep{vovk2005algorithmic} of novel uncertainty quantification algorithms, offering practitioners reliability through specific non-asymptotic guarantees of the type \(\mathbb{P}(Y \in \widehat{C}^{\alpha}(X)) \geq 1 - \alpha\), where $\mathbb{P}$ is the probability law over the random sample $\mathcal{D}_{n}$ and $(X,Y)$. Specifically, we introduce two algorithms tailored to the underlying signal-to-noise ratio of the regression function: homoscedastic and heteroscedastic settings. In both cases, under some regularity conditions, for a fixed point \(x \in \mathcal{X}\), we provide asymptotic conditional consistency results of the form:
    
    \begin{equation}
    \mathbb{P}(Y \in \widehat{C}^{\alpha}(X) \mid X = x) = 1 - \alpha + o_{p}(1).
    \end{equation}
    
    \item We introduce a novel bootstrap approach leveraging the asymptotic Gaussian structure of our estimators to define conditional tolerance predictive regions in probability for random objects in a separable Hilbert space. This extends the seminal work with functional data in the unconditional case by \citet{rathnayake2016tolerance}.
    
    \item We demonstrate the advantages of our novel models across various data structures, including multivariate Euclidean responses, distributional representations with the 2-Wasserstein distance, and standard functional-to-functional data analysis. To the best of our knowledge, our methodology uniquely fills a void in the existing literature by offering a comprehensive framework for quantifying uncertainty within regression models of a functional nature (predictors and response in separable Hilbert spaces), where no other general predictive framework currently exists.
    
    \begin{enumerate}
        \item We adapt our proposed algorithm to handle the survey design of the NHANES study \cite{matabuena2022physical}.
        \item We provide a functional definition of recommended physical exercise levels over the full range of intensities recorded by the accelerometer device. An essential point in the modeling task is that definitions are given from a personalized standpoint, conditioned on factors such as sex, age, and BMI, which dramatically influence daily energy expenditure. This is a significant advancement over existing guidelines that provide standard recommendations about physical activity without conditioning on specific patient characteristics.
    \end{enumerate}
\end{enumerate}

\section{Uncertainty Quantification Algorithms}
\subsection{Background about Tolerance Regions and Depth Bands}

The study of tolerance regions has a long-standing tradition in statistics \cite{li2008multivariate}. For a random variable $Y \sim P$, defining tolerance regions typically involves determining a statistical region $T$ such that

\[
\int_T P(dy) \geq \alpha.
\]

\noindent Here, the region $T$ encompasses at least a fraction $\alpha$ (where $\alpha \in (0,1)$) of the probability mass of $Y$. The modeling goal is to define a tolerance region $T$ that satisfies the nominal coverage level $\alpha$, while also optimizing geometric constraints, such as minimizing the hypervolume of $T$ or another mathematical property.

Originally, tolerance regions were conceptualized under the assumption of Gaussian distributions \cite{murphy1948non, lucagbo2023rectangular}. As the field progressed, methodologies expanded to include semi-parametric \cite{liu2009building} and non-parametric approaches \cite{young2020nonparametric}, primarily dealing with standard multivariate Euclidean data. However, with advancements in technology, especially in fields like medical science, there is now a demand for new methods applicable to random objects such as functions, graphs, or distributions that take values in general metric spaces.

Our research introduces new algorithms for random objects that take values in abstract spaces. Following \cite{li2008multivariate}, we introduce two definitions of tolerance regions for the unconditional case.

\begin{definition}
For $\alpha \in (0,1)$, a random region $T(Y_1, \dots, Y_n)$ is an $\alpha$-content tolerance region (Type I) at confidence level $\gamma \in (0,1)$ if it satisfies:
\[
\mathbb{P}\left( P\left( Y \in T(Y_1, \dots, Y_n) \right) \geq \alpha \right) = \gamma,
\]
where $\mathbb{P}$ denotes the probability over the random sample $\mathcal{D}_n = \{Y_1, \dots, Y_n\}$.
\end{definition}

\begin{definition}
A region $T(Y_1, \dots, Y_n)$ is an $\alpha$-expectation tolerance region (Type II) if it satisfies:
\[
\mathbb{E}\left[ P\left( Y \in T(Y_1, \dots, Y_n) \right) \right] = \alpha,
\]
where the expectation is taken over the random sample $\mathcal{D}_n = \{Y_1, \dots, Y_n\}$.
\end{definition}

\begin{remark}
In both definitions, $P$ denotes the probability distribution of the random variable $Y$, which takes values in an arbitrary separable Hilbert space $\mathcal{Y}$. The parameter $\alpha$ specifies the proportion mass of probability of the population that the tolerance region should cover, while the confidence level $\gamma$ reflects the level of certainty in achieving this coverage.
\end{remark}

The cornerstone of our algorithm is the concept of \textbf{data depth}, a crucial measure for analyzing the structure of data in complex spaces. Consider a random variable \( Y \) in a space \( \mathcal{Y} \) with distribution \( P \). We define data depth as a function \( D(\cdot; P) \) that ranks each point \( y \) in the support of \( Y \) according to its proximity to the geometric center of \( Y \). This center, denoted by \( \overline{y} \), is defined as
\[
\overline{y} = \arg\max_{y \in \mathcal{Y}} D(y; P).
\]
Points near \( \overline{y} \) have higher depth values, reflecting their centrality, whereas points farther from the center have depths approaching zero, indicating they are extreme or outlier points. 

Furthermore, we incorporate an axiomatic topological characterization of data depth, as proposed by \cite{nieto2016topologically}. This framework provides rigorous guidelines for defining and evaluating depth measures, ensuring that they accurately reflect the underlying data's topological and geometrical properties.

\begin{definition}
A statistical depth measure is a mapping \( D: \mathcal{Y} \times \mathcal{P} \to [0,1] \), where \( \mathcal{P} \) is a specified space of probability measures over \( \mathcal{Y} \), that satisfies the following properties:
\begin{enumerate}[i)]
    \item \textbf{ Affine Invariance}: The depth is invariant under affine transformations.
    \item \textbf{ Maximality at the Center}: The depth attains its maximum at the center \( \overline{y} \).
    \item \textbf{ Monotonicity Relative to the Deepest Point}: The depth decreases as one moves away from \( \overline{y} \).
    \item \textbf{ Upper Semi-Continuity}: The depth function \( D(y; P) \) is upper semi-continuous in \( y \in \mathcal{Y} \).
    \item \textbf{ Receptivity to Convex Hull Width}: The depth reflects the spread of the data within its convex hull.
    \item \textbf{ Continuity in \( \mathcal{P} \)}: The depth function is continuous with respect to the probability measure \( P \).
\end{enumerate}
\end{definition}

Complete details regarding the definition are provided in the Supplementary Material.


Data depth, as proposed by \cite{li2008multivariate}, provides a robust framework for establishing unconditional tolerance regions for multivariate data. These methods generalize the notions of order statistics and ranks to multivariate settings.

For simplicity, suppose that the probability distribution \( P \) is continuous. Consider a depth measure \( D(\cdot; P) \) satisfy the before definition. Let \( D(Y_1; P), \dots, D(Y_n; P) \) represent the evaluations of \( n \) independent and identically distributed (i.i.d.) random observations \( \mathcal{D}_n = \{Y_i\}_{i=1}^{n} \) from \( P \). Define \( D^{(i)} \) for \( i = 0, \dots, n \) as the \( i \)-th reverse-order statistic from the set \( \{ D(Y_i; P) \}_{i=1}^{n} \), and set \( D^{(0)} = 1 \) for convenience (see more details in \cite{li2008multivariate}). Here, \( D^{(0)} \) represents the maximal depth value, akin to the geometric median, serving as the most central point under the depth measure.

Based on this ordering, we define different subspaces of \( \mathcal{Y} \) according to the following depth spacings algorithm:

\begin{itemize}
    \item For \( i = 1, \dots, n \),
    \[
    MS_i = \left\{ y \in \mathcal{Y} : D^{(i-1)} > D(y; P) \geq D^{(i)} \right\},
    \]
    with \( D^{(0)} = 1 \).
    \item
    \[
    MS_{n+1} = \left\{ y \in \mathcal{Y} : D(y; P) < D^{(n)} \right\}.
    \]
\end{itemize}

Next, we introduce formal results from \cite{li2008multivariate} that relate the evaluations of depth measures to their probability coverage. 

\begin{theorem}\label{theorem:spacing}
(\cite{li2008multivariate})
Suppose that \( \mathcal{D}_{n} = \{ Y_1, \dots, Y_n \} \) is an i.i.d. sample from \( P \) and \( D(\cdot; P) \) is an affine-invariant depth measure. Then,

\[
\left( P(MS_1), P(MS_2), \dots, P(MS_n) \right) \overset{d}{=} \left( Z_{(1)}, Z_{(2)} - Z_{(1)}, \dots, Z_{(n)} - Z_{(n-1)} \right)
\]

\noindent where \( \mathcal{S}_n = \{ Z_1, \dots, Z_n \} \) is an i.i.d. sample of random variables distributed as \( U[0,1] \), and \( Z_{(i)} \) is the \( i \)-th order statistic of \( \mathcal{S}_n \). 
\end{theorem}

\noindent For each \( r_n \in \{1, 2, \dots, n\} \), we proposes to define the tolerance region \( \widehat{T}^{r_n}(Y_1, \dots, Y_n) \) as the union of a selected number of inner spacings, given by:

\begin{equation}
\widehat{T}^{r_n}(Y_1, \dots, Y_n) = \bigcup_{i=1}^{r_n} MS_i = \left\{ Y : D(Y; P) \geq D^{(r_n)} \right\},
\end{equation}

\noindent where \( MS_i \) represents the \( i \)-th inner spacing defined by the depth measure \( D(Y; P) \) with respect to the distribution \( P \). Drawing on Theorem~\ref{theorem:spacing}, we can infer the distributional properties of the tolerance regions \( \widehat{T}^{r_n}(Y_1, \dots, Y_n) \) for each \( r_n = 1, \dots, n \).

\begin{corollary}
(\cite{li2008multivariate})
Under the conditions of Theorem~\ref{theorem:spacing}, we have
\[
P\left( Y \in \widehat{T}^{r_n}(Y_1, \dots, Y_n) \mid \mathcal{D}_n \right) \overset{d}{=}  \textnormal{Beta}(r_n, n + 1 - r_n), \quad \text{for } r_n = 1, \dots, n.
\]
\end{corollary}

\begin{remark}
In the case of constructing an \( \alpha \)-expectation tolerance region, we use the property that \( \mathbb{E}\left[ \text{Beta}(r_n, n + 1 - r_n) \right] = \frac{r_n}{n + 1} \). Thus, to achieve a nominal coverage level \( \alpha \), we set \( r_n = \lceil (n + 1) \alpha \rceil \).
\end{remark}

\noindent \subsection{Kernel Mean Embedding and Integrated Depth Bands}
   
   This paper exploits the concept of depth bands to derive prediction regions. Particularly, we  focused on a special type on $h$-integrated depth band \citep{mosler2013depth, CuevasFF07}, which mathematically corresponds to the notion of kernel mean embedding \citep{wynne2021statistical}. Kernel mean embeddings \citep{muandet2017kernel} are based on the idea of mapping probability distributions into a reproducing kernel Hilbert space (RKHS), allowing the application of the full arsenal of kernel methods to probability measures. They can be viewed as a generalization of the original ``feature map'' common to support vector machines (SVMs) and other kernel methods.

For any positive definite kernel function $k_{\mathcal{Y}}: \mathcal{Y}\times \mathcal{Y}\to \mathbb{R}^{+}$, there exists a unique reproducing kernel Hilbert space (RKHS) $\mathcal{H}_{\mathcal{Y}}$. The RKHS $\mathcal{H}_{\mathcal{Y}}$ is a (possibly infinite-dimensional) space of functions $h: \mathcal{Y} \rightarrow \mathbb{R}$ where evaluation can be expressed as an inner product: for all $h, f \in \mathcal{H}_{\mathcal{Y}}$, we have $h(f) = \langle h, k_{\mathcal{Y}}(\cdot, f) \rangle_{\mathcal{H}_{\mathcal{Y}}}$. 


For a given marginal distribution $P_Y$, the kernel mean embedding (KME) $\mu_{P_Y}$ is defined as the expectation of the feature map $k_{\mathcal{Y}}(\cdot, Y)$:
\[
\mu_{P_Y} = \mathbb{E}_{Y}[k_{\mathcal{Y}}(\cdot, Y)] \in \mathcal{H}_{\mathcal{Y}}.
\]

The KME always exists for bounded kernels. Furthermore, for characteristic kernels, these embeddings are injective, uniquely determining the probability distribution $P_Y$ \cite{sriperumbudur2008injective, sriperumbudur2010hilbert}. Popular kernels like Gaussian and Laplace kernels possess this property.

The second-order mean embeddings, also termed covariance operators in \cite{fukumizu2004dimensionality}, are defined as the expectation of tensor products between features:
\[
C_{XY} = \mathbb{E}_{XY}[k_{\mathcal{X}}(\cdot, X) \otimes k_{\mathcal{Y}}(\cdot, Y)],
\]
\noindent where $\otimes$ denotes the tensor product.\footnote{We introduce the notation $\otimes$ to represent the tensor product between elements of RKHSs. The tensor product $k_{\mathcal{X}}(\cdot, X) \otimes k_{\mathcal{Y}}(\cdot, Y)$ is an element of the tensor product space $\mathcal{H}_{\mathcal{X}} \otimes \mathcal{H}_{\mathcal{Y}}$.}The covariance operators generalize the familiar notion of covariance matrices to infinite-dimensional feature spaces and always exist for bounded kernels.

For a set of i.i.d. samples $Y_1, \ldots, Y_n$, the kernel mean embedding can be estimated by its empirical version:
\begin{equation}
\widehat{\mu}_{P_Y} = \frac{1}{n} \sum_{i=1}^{n} k_{\mathcal{Y}}(\cdot, Y_i).
\end{equation}

From this estimator, various associated quantities, including estimators of squared RKHS distances \citep{gretton2012kernel} between embeddings needed for kernel-based hypothesis tests, can be derived.

From a mathematical statistics perspective, kernel mean embeddings can be considered a special case of $h$-depths. Explicit mathematical connections and relationships are provided in the Supplemental Material $S_1$ (see the complete description in Theorem 3 from \cite{wynne2021statistical}).

Given this connection, the empirical estimator $\widehat{\mu}_{P_Y}$ can be considered in practice as a data depth measure $\widehat{D}_{k}(\cdot; P_Y)$. Thus, we have:
\[
\widehat{D}_{k}(\cdot; P_Y) = \widehat{\mu}_{P_Y} = \frac{1}{n} \sum_{i=1}^{n} k_{\mathcal{Y}}(\cdot, Y_i).
\]
This estimator $\widehat{D}_{k}(\cdot; P_Y)$ can be utilized within our framework to obtain $\alpha$-expectation tolerance regions $\widehat{T}(Y_1, \dots, Y_n)$ using Algorithm~\ref{alg:alpha_expectation_tolerance}.

\begin{algorithm}[H]
\caption{Estimation of $\alpha$-Expectation Tolerance Regions}
\label{alg:alpha_expectation_tolerance}
\begin{algorithmic}[1]
  \State \textbf{Input:} Dataset $\{Y_j\}_{j=1}^{n}$, confidence level $\gamma \in (0,1)$, tolerance level $\alpha \in (0,1)$.
  
  \State \textbf{Output:} $\alpha$-expectation tolerance region $\widehat{T}(Y_1,\dots,Y_n)$.
  
  \State Estimate the kernel-based depth for each observation $Y_j$ in the dataset $\{Y_j\}_{j=1}^{n}$:
    \[
    \widehat{D}_{k}(Y_j; P_{Y}) = \frac{1}{n} \sum_{i=1}^{n} k_{\mathcal{Y}}(Y_j, Y_i),
    \]
  \noindent where $k_{\mathcal{Y}}(\cdot,\cdot)$ denotes the kernel function. 
  \State Sort the estimated depths $\{\widehat{D}_{k}(Y_j; P_{Y})\}_{j=1}^{n}$ in non-increasing order to obtain the order statistics:
    \[
    \widehat{D}^{(1)} \geq \widehat{D}^{(2)} \geq \cdots \geq \widehat{D}^{(n)}.
    \]
    
  \State Define the $\alpha$-expectation tolerance region $\widehat{T}(Y_1, \dots, Y_n)$ as the set of points $Y$ satisfying:
    \[
    \widehat{T}(Y_1, \dots, Y_n) = \left\{ Y : \widehat{D}_{k}(Y; P_{Y}) \geq \widehat{D}^{\left\lceil n + 1 - \alpha(n + 1) \right\rceil} \right\},
    \]
    where $\left\lceil \cdot \right\rceil$ denotes the ceiling function, ensuring that the index is an integer.
\end{algorithmic}
\end{algorithm}

\subsection{Conditional Conformal Depth Bands Algorithm with Non-Asymptotic Guarantees}

In this section, we focus on using the theory of tolerance regions to propose an explicit prediction region algorithm that incorporates conditional kernel mean embedding (for estimation details see \cite{park2020measure}). We replace the notation of the $\alpha$-tolerance region $T(Y_1, \dots, Y_n)$ with $\widehat{\mathcal{C}}^\alpha(\cdot)$. Additionally, we define the population prediction region as

\begin{equation}
C^\alpha(x) = \left\{ y \in \mathcal{Y} :  g(x,y):= P(D_k(Y; P_{Y\mid X}) \leq  D_k(y; P_{Y\mid X=x})  \mid  X=x) \geq q_{1-\alpha}(x) \right\},
\end{equation}

\noindent where $q_{1-\alpha}(x)$ is the $(1-\alpha)$ quantile of the conditional probability distribution of the depth values for all $x\in \mathcal{X}$ given $X = x$.

For a population perspective, under the continuity hypothesis and due to the properties of depth measures and integral transformation ranks, we have \( q_{1-\alpha}(x) = 1 - \alpha \). 

Algorithm \ref{algorithm:pred} presents the general formulation. The high‑level idea is as follows: first, estimate the conditional kernel mean embedding \(\widehat{D}_k(\cdot; P_{Y\mid X_i})\); next, project the observations from the held‑out split according to the ranking induced by this embedding; and finally, define the conditional level set by inverting the corresponding distribution function.

\begin{algorithm}
\caption{General Prediction Region Algorithm}
\label{algorithm:pred}
\begin{algorithmic}[1]
    \Require Dataset $\mathcal{D}_n = \{(X_i, Y_i)\}_{i=1}^{n}$, confidence level $\alpha \in (0,1)$

\Ensure Prediction region
\[
\widehat{C}^{\alpha}(x)
= \bigl\{\,y \in \mathcal{Y} : \widehat{g}(x,y) \ge \widehat{q}_{1-\alpha}\bigr\},
\]
\noindent where \(\widehat{g}(x,y)\) is an estimator of \(g(x,y)\) (see Equation $5$).

    \State Split $\mathcal{D}_n$ into three disjoint sets: $\mathcal{D}_{\text{train}}$, $\mathcal{D}_{\text{calibration}}$, and $\mathcal{D}_{\text{test}}$
    \State Use $\mathcal{D}_{\text{train}}$ to estimate the conditional depth measure $\widehat{D}_k(y; P_{Y|X=x})$  according to \cite{park2020measure}.
    
    \ForAll{$(X_i, Y_i) \in \mathcal{D}_{\text{calibration}}$}
        \State Compute $r_i = \widehat{D}_k(Y_i; P_{Y|X_i})$
    \EndFor
    
    \State Estimate the conditional CDF $\widehat{g}(x, r)$ of the depth values using $\{(X_i, r_i)\}_{i \in \mathcal{D}_{\text{calibration}}}$
    
    \ForAll{$(X_i, Y_i) \in \mathcal{D}_{\text{test}}$}
        \State Compute $s_i = \widehat{g}(X_i, \widehat{D}_k(Y_i; P_{Y|X_i}) )$
    \EndFor
    
    \State Calculate the empirical $(1 - \alpha)$ quantile $\widehat{q}_{1 - \alpha}$ from the values $\{ s_i \}_{i \in \mathcal{D}_{\text{test}}}$
    
    \State \Return $\widehat{C}^{\alpha}(x) = \left\{ y \in \mathcal{Y} : \widehat{g}(x, y) \geq \widehat{q}_{1 - \alpha} \right\}$
\end{algorithmic}
\end{algorithm}

In the homoscedastic case, for any \( x, x' \in \mathcal{X} \), we have \( g(x, y) = g(x', y) \) for all \( y \in \mathcal{Y} \), because the depth is affine--invariant and we assume a scale and localization model. For simplicity, we denote \( g(y) := g(\cdot, y) \). To adapt Algorithm~\ref{algorithm:pred}, we estimate the conditional distribution of the depth-band measure using the empirical distribution derived from the random elements \( \{ r_i \}_{i \in \mathcal{D}_{\text{test}}} \). As a result, we no longer require three splits in the uncertainty quantification procedure. Algorithm~\ref{algorithm:homoscedastic} outlines the core steps of the proposed method for this setting.

\begin{algorithm}[H]
\caption{Prediction Region Algorithm for the Homoscedastic Case}
\label{algorithm:homoscedastic}
\begin{algorithmic}[1]
    \State Split $\mathcal{D}_n$ into two disjoint sets: $\mathcal{D}_{\text{train}}$ and $\mathcal{D}_{\text{test}}$.
    \State Use $\mathcal{D}_{\text{train}}$ to estimate the conditional depth measure $\widehat{D}_k(y; P_{Y|X})$.
    \State Define $D(x, y) = \widehat{D}_k(y; P_{Y|X})$ for all $x \in \mathcal{X}$ and $y \in \mathcal{Y}$.
    \State For each $(X_i, Y_i) \in \mathcal{D}_{\text{test}}$, compute $r_i = D(X_i, Y_i)$.
    \State Calculate the empirical $(1 - \alpha)$ quantile $\widehat{q}_{1 - \alpha}$ from the sample $\{ r_i \}_{i \in \mathcal{D}_{\text{test}} }$.
    \State \textbf{Return} the prediction region $\widehat{C}^{\alpha}(x) = \{ y \in \mathcal{Y} : D(x, y) \geq \widehat{q}_{1 - \alpha} \}$.
\end{algorithmic}
\end{algorithm}

\begin{assumption}\label{assumption:homoscedastic}
Suppose that the following hold:
\begin{enumerate}
    \item $n_{\text{train}},\ n_{\text{test}} \to \infty$ as $n \to \infty$.
    \item The estimator $\hat{\mu}_{P(Y \mid X)}$ is consistent in the sense that
    \[
    \lim_{n_{\text{train}} \to \infty}  \left\| \hat{\mu}_{P(Y \mid X)} - \mu_{P(Y \mid X)} \right\|_{\infty} = 0, \quad \text{in probability}.
    \]
    \item The population quantile \( q_{1 - \alpha} = \inf\{ t \in \mathbb{R} : G(t) = P( D_k(Y; P_{Y|X}) \geq t ) = 1 - \alpha \} \) exists uniquely and is a continuity point of the function \( G(\cdot) \). By definition, under continuty hyphotesis, $q_{1 - \alpha} = 1 - \alpha$.
\end{enumerate}
\end{assumption}

\begin{theorem}\label{theorem:homoscedastic_convergence}
Under Assumption~\ref{assumption:homoscedastic}, the estimated prediction region $\widehat{C}^{\alpha}(x; \mathcal{D}_{n})$ obtained using Algorithm~\ref{algorithm:homoscedastic} satisfies
\[
\int_{\mathcal{X}} \mathbb{P}\left( Y \in \widehat{C}^{\alpha}(X) \triangle C^{\alpha}(X) \mid X = x, \mathcal{D}_{n} \right) P_{X}(dx) = o_p(1),
\]
\noindent where \(C^{\alpha}(x)\) is the true prediction region at level \(\alpha\), \(P_X\) denotes the probability law of \(X\), and \(\triangle\) represents the symmetric difference between sets. In particular, for any sets \(A\) and \(B\), 
$A \triangle B = (A \setminus B) \cup (B \setminus A).$
\end{theorem}
\subsection{Heteroscedastic Case}\label{sec:heteroscedastic}
In the \emph{heteroscedastic} setting, we seek parsimonious, interpretable
models for the conditional distribution
\(g(x,y)=P(D_k(Y; P_{Y\mid X}) \leq  D_k(y; P_{Y\mid X=x})  \mid  X=x)\).
To this end, we adopt the semi-parametric \emph{generalized additive models
for location, scale, and shape} (GAMLSS) framework introduced by
\citet{rigby2005generalized}.  GAMLSS extends classical mean-regression by
allowing \emph{all} distributional parameters to depend on covariates, thus
accommodating skewed or heavy-tailed responses often encountered in
practice.

Alternative fully non-parametric estimators are available—for example the
$k$-nearest-neighbour (kNN) conditional distribution estimator
\citep{dombry2023stone,matabuena2024knn} or the
Nadaraya–Watson kernel estimator
However, their mean-squared error deteriorates quickly with the dimension
\(d=\dim(X)\), whereas GAMLSS retains root-\(n\) rates while remaining
flexible.

Let \(Y\mid X=x\sim F\bigl(\mu(x),\sigma(x),\nu(x),\tau(x)\bigr)\),
where
\(\boldsymbol{\theta}(x)=\bigl(\mu,\sigma,\nu,\tau\bigr)^{\!\top}\)
controls location, scale, skewness, and kurtosis (some components may be
omitted when unnecessary). For each parameter component
\(\theta_{\ell}(x)\) (\(\ell=1,\dots,L\), \(L\le4\)) we posit the additive
predictor
\[
  g_{\ell}\!\bigl\{\theta_{\ell}(x)\bigr\}
    \;=\;
    \eta_{\ell}(x)
    \;=\;
    \mathbf{x}_{\ell}(x)^{\!\top}\boldsymbol{\beta}_{\ell}
    \;+\;
    \sum_{j=1}^{J_{\ell}}
        z_{\ell j}(x)^{\!\top}\boldsymbol{\gamma}_{\ell j},
    \tag{1}
\]
\noindent where \(g_{\ell}\colon\mathbb{R}\to\mathbb{R}\) is a known monotone link;
\(\mathbf{x}_{\ell}(x)\in\mathbb{R}^{p_{\ell}}\) collects fixed-effect
        basis functions (e.g.\ splines);  
\(\boldsymbol{\beta}_{\ell}\in\mathbb{R}^{p_{\ell}}\) are
        corresponding coefficients;
\(z_{\ell j}(x)\in\mathbb{R}^{q_{\ell j}}\) are random-effect bases,
        with
        \(\boldsymbol{\gamma}_{\ell j}\sim
          \mathcal{N}_{q_{\ell j}}(\mathbf 0,G_{\ell j})\).

Stacking the evaluations over the sample
\(\{(X_i,Y_i)\}_{i=1}^{n}\) yields design matrices
\(\mathbf{X}_{\ell}\in\mathbb{R}^{n\times p_{\ell}}\) and
\(\mathbf{Z}_{\ell j}\in\mathbb{R}^{n\times q_{\ell j}}\).
Parameter estimation proceeds via (penalised) maximum likelihood; the
resulting M-estimators are consistent and asymptotically normal under
standard regularity conditions \cite{van1996weak}.

\textbf{Empirical-process notation.}
Let \(Y_1,\dots,Y_n\) be i.i.d.\ with law \(P\) on
\((\mathcal Y,\mathcal A)\). For a measurable
\(f\colon\mathcal Y\to\mathbb R\) define
\[
   Pf=\int f\,\mathrm dP,\qquad
   \mathbb P_n f=\frac1n\sum_{i=1}^{n}f(Y_i),\qquad
   \mathbb G_n f=\sqrt n\,(\mathbb P_n-P)f.
\]

Consider a function class
\(\{f_{\theta,\eta}:\theta\in\Theta,\eta\in H\}\)
and data-dependent ``estimators'' \(\eta_n\in H\).
Following \citet{van2007empirical} we aim to show
\begin{equation}\label{eqnw}
  \sup_{\theta\in\Theta}
  \bigl|\mathbb G_n\{f_{\theta,\eta_n}-f_{\theta,\eta_0}\}\bigr|
  \;\xrightarrow{p}\;0,
  \tag{2}
\end{equation}
\noindent where \(\eta_0\) is the probability limit of \(\eta_n\).
If (2) holds  and
\(\eta_n\) is $P$-consistent,
\(\sup_{\theta}P\{f_{\theta,\eta_n}-f_{\theta,\eta_0}\}^2\to0\),
then the empirical process
\(\mathbb G_n f_{\theta,\eta_0}\) admits a
Gaussian weak limit, and standard delta-method arguments yield the
asymptotic distribution of \(\mathbb P_n f_{\theta,\eta_n}\).

\begin{proposition}[{\citealp{van2007empirical}}]
Let \(H_0\subset H\) satisfy
\(\Pr\{\eta_n\in H_0\}\to1\) and assume that the class
\(\{f_{\theta,\eta}:\theta\in\Theta,\eta\in H_0\}\) is $P$-Donsker.
Then \ref{eqnw} is valid.
\end{proposition}

\begin{proposition}[Uniform consistency of $\widehat g$]
Assume \(\mathcal X\subset\mathbb R^{d}\) is finite-dimensional and the
kernel mean-embedding class is $P$-Donsker.  If $g$ is estimated via a
GAMLSS model fitted by maximum likelihood, then
\[
   \sup_{(x,y)\in\mathcal X\times\mathcal Y}
        \bigl|\,\widehat g(x,y)-g(x,y)\bigr|
        =o_p(1),\qquad
   \sqrt n\bigl\{\widehat g-g\bigr\}
        \;\Rightarrow\;Z,
\]
where \(Z\) is a tight centered Gaussian process.
\end{proposition}

\begin{remark}
Recent work on the Donsker property of kernel mean embeddings
\citep{carcamo2024uniform,pmlr-v201-park23a}
shows that when responses lie in infinite-dimensional
spaces the classical hypothesis may fail; nonetheless, bootstrap
procedures remain valid under weaker conditions
\citep{martinez2024efficient}.
\end{remark}

\subsection{Tolerance Regions in Probability via Bootstrapping}

The literature of tolerance region with functional data or more general, for random objects in metric spaces is scarce. For example, \cite{rathnayake2016tolerance} propose an unconditional tolerance region in probability in Hilbert spaces using functional principal component analysis. However, the authors restrict their analysis to the geometry of the supremum norm  and perform the calibration of the estimator with bootstrap following prior work by \cite{goldsmith2013corrected}. Data depth provides new opportunities to create more general algorithms from both conditional and unconditional perspectives, for instance, using kernel mean embeddings beyond functional data as in the case of spaces of random graphs.

For a fixed $x \in \mathcal{X}$, we define the notion of a conditional tolerance region in probability.

\begin{definition}
For a fixed $X = x$, a random region $T_{x}(Y_1, \dots, Y_n)$ is an $\alpha$-content tolerance region (Type I) at confidence level $\gamma \in (0,1)$ if it satisfies:
  \begin{equation*}
    \mathbb{P}\left( P\left( Y \in T_{x}(Y_1, \dots, Y_n) \mid X = x \right) \geq \alpha \,\bigg|\, \mathcal{D}_n \right) = \gamma,
  \end{equation*}
\end{definition}

\noindent where $\mathbb{P}$ is the joint probability distribution of the random sample $\mathcal{D}_n$.

The basic idea of our algorithm is to exploit the asymptotic Gaussianity of the conditional distribution estimators $g(x, y)$ indexed by a kernel mean embedding. We combine this with Efron's bootstrap \cite{efron1987better} to obtain tolerance regions in probability. Algorithm~\ref{algorithm:tol} outlines the core steps of our procedure.  
We generate \(B\) bootstrap estimators by drawing samples  
\(\mathcal{D}_n^{b}\) with replacement from the original data  
\(\mathcal{D}_n\), for \(b = 1,\dots,B\).  
Each bootstrap estimator is designed to approximate the conditional probability  
\[
  P\!\bigl\{\,Y \in T_x\!\bigl(Y_1,\dots,Y_n\bigr) \,\bigm|\, X = x\bigr\} 
  \;\ge\; \alpha .
\]  
Conditional on \(\mathcal{D}_n\), the bootstrap datasets  
\(\mathcal{D}_n^{b}\) are i.i.d.  
We then calibrate the tolerance level \(\gamma\) using the empirical  
\((1-\alpha)\)-quantile, denoted \(\widehat{q}_{1-\alpha}\), computed from the  
\(B\) bootstrap estimates.


\begin{algorithm}[ht!]
\caption{Conditional Tolerance Region in Probability}
\label{algorithm:tol}
\textbf{Input:} Dataset $\mathcal{D}_n = \{(X_i, Y_i)\}_{i=1}^{n}$, confidence level $\alpha \in (0,1)$, desired tolerance level $\gamma \in (0,1)$, a point $x \in \mathcal{X}$, number of bootstrap samples $B$.\\
\textbf{Output:} Tolerance region in probability $\widehat{C}^{\text{tolerance}, \alpha}(x) = \left\{ y \in \mathcal{Y} : \widehat{g}(x, y) \geq \widehat{q}^{\gamma}_{1 - \alpha} \right\}$.

\begin{algorithmic}[1]
    \For{$b = 1$ to $B$}
        \State Sample with replacement $n$ observations from $\{1, 2, \dots, n\}$ to obtain indices $\mathcal{I}_b$.
        \State Define the bootstrap dataset $\mathcal{D}_n^b = \{ (X_i, Y_i) \}_{i \in \mathcal{I}_b}$.
        \State Using $\mathcal{D}_n^b$, estimate the prediction region at confidence level $1 - \alpha$ and obtain the quantile $\widehat{q}_{1 - \alpha}^b$.
    \EndFor
    \State Define the empirical distribution function of the bootstrap quantiles:
    \[
    \widehat{F}_B^q(t) = \frac{1}{B} \sum_{b=1}^B \mathbf{1}\left\{ \widehat{q}_{1 - \alpha}^b \leq t \right\}.
    \]
    \State Obtain the adjusted quantile $\widehat{q}^{\gamma}_{1 - \alpha}$ such that $\widehat{F}_B^q\left( \widehat{q}^{\gamma}_{1 - \alpha} \right) = \gamma$.
    \State \textbf{Return} the tolerance region in probability:
    \[
    \widehat{C}^{\text{tolerance}, \alpha}(x) = \left\{ y \in \mathcal{Y} : \widehat{g}(x, y) \geq \widehat{q}^{\gamma}_{1 - \alpha} \right\}.
    \]
\end{algorithmic}
\end{algorithm}

\section{Simulation Study}
The goal of this Simulation section is to demonstrate the strength of our methodology in finite--sample regimes relative to well--established conformal approaches. The analysis is divided into two parts. (i) In a scalar--to--scalar setting, we evaluate conditional coverage and compare the results with state--of--the--art methods designed for this scenario. (ii) In a functional--to--functional setting--where, to the best of our knowledge, no alternative conformal methods are currently available--we assess our model’s performance in terms of marginal coverage. Estimating conditional coverage for multivariate or functional predictors with non--parametric tools is generally unreliable unless the sample size is very large; even then, uncertainty quantification requires thousands of Monte Carlo replications, which is computationally demanding. Consequently, we restrict the conditional--coverage study to the univariate case, where reliable benchmarks exist. For the functional example, we concentrate on marginal coverage, which is more appropriate given the complexity of the output space and the absence of established competitors. Additional simulations for other data structures, together with marginal--coverage results, are presented in the Supplementary Material.



\subsection{Conditional Coverage Analysis}

\begin{table}[ht!]
\centering
\caption{Mean–squared conditional–coverage errors for Settings~1 and~2.}
\small            
\setlength{\arraycolsep}{4pt}  

\scalebox{0.85}{%
\(
\renewcommand{\arraystretch}{1.15}
\begin{array}{lcccccc|cccccc}
\hline
& \multicolumn{6}{c|}{\textbf{Setting 1}}
& \multicolumn{6}{c}{\textbf{Setting 2}}\\
\cline{2-13}
& \multicolumn{2}{c}{\alpha = 0.80}
& \multicolumn{2}{c}{\alpha = 0.90}
& \multicolumn{2}{c|}{\alpha = 0.95}
& \multicolumn{2}{c}{\alpha = 0.80}
& \multicolumn{2}{c}{\alpha = 0.90}
& \multicolumn{2}{c}{\alpha = 0.95}\\
\cline{2-13}
\text{Method}
& n{=}1000 & n{=}3000 & n{=}1000 & n{=}3000 & n{=}1000 & n{=}3000
& n{=}1000 & n{=}3000 & n{=}1000 & n{=}3000 & n{=}1000 & n{=}3000\\
\hline
\text{Quantile–CQR}
& 0.454 & 0.453 & 0.341 & 0.338 & 0.179 & 0.177
& 0.0211 & 0.0194 & 0.0215 & 0.0193 & 0.0132 & 0.0133\\
\text{HPD–1}
& 0.218 & 0.215 & 0.101 & 0.099 & 0.029 & 0.030
& 0.0526 & 0.0537 & 0.0256 & 0.0269 & 0.0122 & 0.0100\\
\text{HPD–2}
& 0.222 & 0.224 & 0.104 & 0.105 & 0.050 & 0.050
& 0.0465 & 0.0478 & 0.0286 & 0.0294 & 0.0143 & 0.0134\\
\textbf{Proposed}
& \mathbf{0.031} & \mathbf{0.034} & \mathbf{0.026} & \mathbf{0.029} & \mathbf{0.011} & \mathbf{0.012}
& \mathbf{0.0484} & \mathbf{0.0488} & \mathbf{0.0291} & \mathbf{0.0303} & \mathbf{0.0210} & \mathbf{0.0175}\\
\hline
\end{array}
\)
} 
\label{tab:joined_settings}
\end{table}

We evaluate the conditional coverage
\[
  P\!\bigl( Y \in C^{\alpha}(X) \mid X \bigr)
\]
of our heteroscedastic procedure and compare it with established conformal methods for Euclidean data: the HPD–split scores proposed by \citet{izbicki2022cd} and two variants of Conformalized Quantile Regression (CQR) \citep{romano2019conformalized, sesia2020comparison}.  
The HPD–split method is implemented with the \texttt{dist} and \texttt{hpd} functions available at \url{https://github.com/rizbicki/predictionBands}, while CQR is executed with the default settings of the \texttt{cfsurvival} package.

\textbf{Monte Carlo design.}  
For each setting, sample size $n\in \{1000,3000\},\alpha\in \{0.8,0.9,0.95\},$ we generate \(B = 100\) Monte Carlo data sets of size \(n_4 = 2000\).  
In each replicate, we fit a generalized additive model to the indicator variables \(\mathbb{I}\{Y \in \widehat{C}^{\alpha}(X)\}\) in order to estimate
\[
  \widehat{p}^{\alpha}(x)
  = \mathbb{P}\!\bigl( \mathbb{I}\{Y \in \widehat{C}^{\alpha}(x)\} = 1 \mid X = x \bigr).
\]
Deviation from the nominal level \(\alpha\) is summarized, for each replicate \(b \in \{1,\dots,B\}\), by the \(L^{2}\) error
\[
  \widehat{\mathrm{Err}}_{b}^{\alpha}
  = \int_{a}^{b} \bigl[ \widehat{p}^{\alpha}(x) - \alpha \bigr]^{2}\,dx,
\]
where \([a,b]\) denotes the support of \(X\).  
The results reported in Table~\ref{tab:joined_settings} correspond to the average
\[
  \frac{1}{B}\sum_{b=1}^{B}\widehat{\mathrm{Err}}_{b}^{\alpha}.
\]

\textbf{Data-generating mechanisms.}  
We first consider a nonlinear heteroscedastic setting:
\[
  X \sim \mathcal{U}(0,5), 
  \quad
  \varepsilon \sim \mathcal{U}(-1,1), 
  \quad
  Y = 3 + \exp(X) + \varepsilon\,X,
\]
and then a linear homoscedastic setting:
\[
  X \sim \mathcal{U}(0,5), 
  \quad
  \varepsilon \sim \mathcal{U}(0,5), 
  \quad
  Y = X + \varepsilon.
\]

\textbf{Results.}  
In the nonlinear heteroscedastic scenario, our method clearly outperforms the competitors in terms of conditional coverage accuracy.  
In the simple homoscedastic scenario, all methods perform comparably well, so model choice is less critical.  
Note, however, that our approach is tailored for heteroscedasticity and may be suboptimal in a homoscedastic regime.

\subsection{Functional-to-Functional Regression}

To illustrate the versatility of our uncertainty-quantification algorithm, we consider a canonical functional-to-functional regression model with predictor and response in $L^{2}([0,1])$. To the best of our knowledge, no competing methods for uncertainty quantification in this setting have yet been proposed.

We assume the regression relationship
\[
  Y(t) \;=\; \beta_{0}(t) \;+\; \int_{0}^{1} X(s)\,\beta(s,t)\,ds \;+\; \varepsilon(t),
\]
\noindent where the covariate $X(\cdot)\in L^{2}([0,1])$, the response $Y(\cdot)\in L^{2}([0,1])$, and the slope surface $\beta(\cdot,\cdot)\in L^{2}([0,1]^2)$. The data-generating components are specified as follows:
\begin{align*}
  \beta_{0}(t) &= \sin(\pi t), &
  \beta(s,t) &= 5st, &
  X(s) &= s\eta,\quad \eta \sim \mathcal{N}(0,\sigma_\eta^{2}),\\
  \varepsilon(t) &= \cos(2\pi t)\,\varepsilon_{1} + \sin(2\pi t)\,\varepsilon_{2}, &
  & \varepsilon_{j}\sim \mathcal{N}(0,\sigma_{j}^{2}),\; \sigma_{1}=0.5,\; \sigma_{2}=0.75.
\end{align*}
\noindent Each realization of the predictor is generated as
\[
  X_{i}(s)=\sum_{k=1}^{10}\psi_{ik}\,\phi_{k}(s), \qquad
  \psi_{ik}\overset{\text{i.i.d.}}{\sim}\mathcal{N}\!\bigl(0,\sigma_{k}^{2}\bigr),\;
  \sigma_{k}^{2}=10-k+1,
\]
\noindent where $\{\phi_{k}\}_{k=1}^{10}$ denotes an orthonormal polynomial basis on $[0,1]$.

\begin{figure}[ht!]
\begin{center}
\begin{tabular}{ll}
\includegraphics[width=.5\linewidth , height=.5\linewidth]{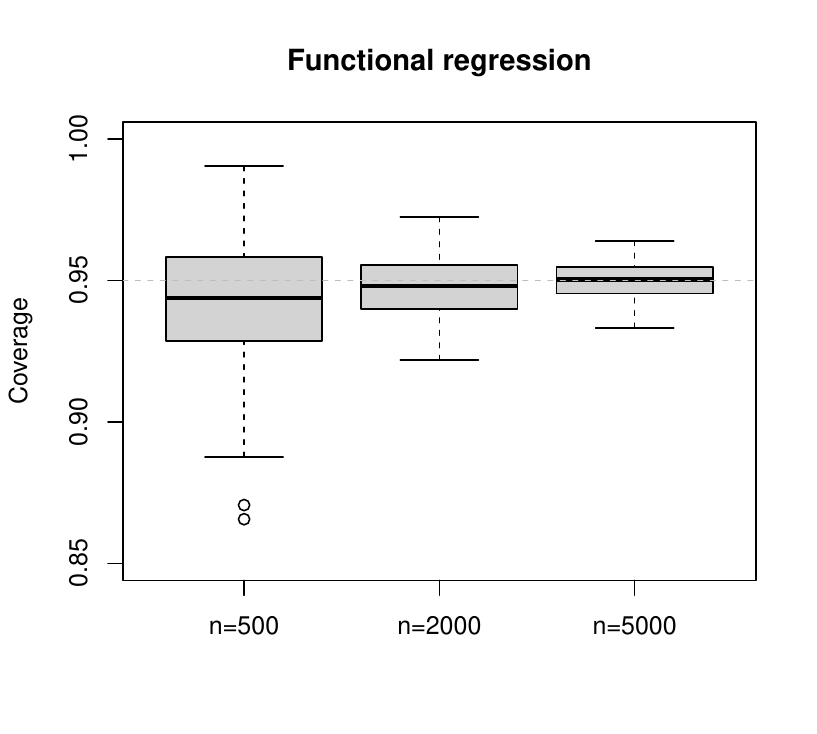} &
\end{tabular}
\end{center}
\caption{Estimated marginal coverage, functional to functional regression example, $\alpha=0.05$.}
\label{fig:figfunctional}
\end{figure}

Figure~\ref{fig:figfunctional}\,(left) shows the empirical marginal coverage at the $1-\alpha = 0.95$ level for sample sizes $n\in\{500,\,2000,\,5000\}$. The results confirm that the finite-sample validity of our method is preserved and that, as $n$ increases, the empirical coverage converges to the nominal level.


\section{Clinical Case Study: Physical Activity Recommendations with NHANES Accelerometer Data}

\subsection{Motivation and Data Description}

Physical exercise is considered one of the most effective pharmacological interventions across a broad spectrum of diseases, as well as a therapy for reversing physiological decline with age or improving overall health \cite{pedersen2006evidence,almeida2014150,bolin2018physical,mendes2016exercise,franks2017causal,friedenreich2021physical,burtscher2020run}. Although many medical guidelines suggest standard recommendations, such as performing $150$ minutes of aerobic exercise per week, there is a growing consensus in the community about the need for a personalized prescription, as well as patient assessment, necessary to ensure the success of the intervention performed \citep{mendes2016exercise,matabuena2019prediction,buford2013toward}. Nowadays, with the modern accelerometer and smartphones, we have more reliable ways to characterize physical activity levels of an individual.

Data modeling is a critical step in accelerometer data analysis for exploiting the rich information that accelerometers capture about physical activity patterns at different resolution time scales. However, as subjects are usually monitored in free-living conditions, practitioners usually resort to basic summary statistics, such as total expenditure normalized by the time the device is worn or compositional metrics. In our recent work \cite{10.1093/jrsssc/qlad007}, we have proposed a new functional distributional profile of the physical activity performed by the individual. The new representation automatically captures the previous summary metrics. Additionally, in the same paper, we generalize a series of nonparametric regression models to incorporate the complex survey design of the NHANES cohort and include kernel ridge regression methods to demonstrate the advantages of the new representation over traditional summary measures of accelerometer data.

The NHANES data provide new scientific opportunities for characterizing physical activity patterns across different age groups, ethnicity, or disease sub-types due to their robust and randomization study design in the American Population, unlike another important physical activity cohort such as the US population. In this study, we select healthy individuals aged 18-80 years and exclude patients with specific comorbidities and other reasonable exclusion criteria explained in detail in the Supplementary Material.

The goal of this paper is to use the new conditional prediction regions algorithm to define, according to specific characteristics of the patients, such as age, sex, BMI, what is the expected distributional representation of the physical activity. Roughly speaking, we can define personalized physical activity recommendations for the individual of the US population. In this way, it provides a sophisticated tool to determine when physical activity interventions patterns must be carried out.

Given the increase in the use of wearable technology to monitor health status at the population level, the methods introduced here have the potential to contribute to public health policies on physical activity traditionally designed for the average person. 

\subsection{Modeling Accelerometer Data with Distributional Data}\label{section:representacion}
We begin by introducing the formal definition of the accelerometer‐based distributional representation, following the framework of Matabuena and Petersen (2023)\cite{10.1093/jrsssc/qlad007}.  For each subject \(i\), let \(n_i\) denote the total number of observations, recorded as pairs \((t_{ij}, A_{ij})\) for \(j=1,\dots,n_i\).  Here, \(t_{ij}\in[0,T_i]\) is the time at which the accelerometer records a physical‐activity measurement \(A_{ij}\).  To account for periods of inactivity, we place a point mass at zero whose weight equals the fraction of time the device registers no movement.  Because the range of \(A_{ij}\) can differ substantially across individuals and populations, standard compositional‐data approaches may struggle; see, for example, Bayes–Hilbert methods or transformation‐based functional techniques.

To address these challenges, we propose using a cumulative distribution function $F_i(t)$ for each individual. Let $Y_i(t)$ be a latent process such that the accelerometer measures $A_{ij} = Y_i(t_{ij})$ $(j=1,\ldots,n_i)$, and define $F_i$ as

\begin{equation}\label{rep1}
F_i(x) = \frac{1}{T_i} \int_{0}^{T_i} \mathbf{1}\{Y_i(t) \leq x\} \, dt, \quad \text{for } x \geq 0.
\end{equation}

In this paper, we consider an invariant positive kernel $k: \mathcal{Y}\times \mathcal{Y}\to \mathbb{R}^{+}$ equipped with the $2$-Wasserstein distance. Specifically, we define $k(F,G)=\kappa(d_{W^{2}}\left(F,G\right))$, where $\kappa(\sqrt{\cdot})$ is a completely monotone, positive function. In our data analysis examples, we use an explicit $2$-Wasserstein Gaussian kernel defined as

\begin{equation}
k(F,G)= e^{ \sigma \int_{0}^{1} (F^{-1}(s)-G^{-1}(s))^{2} \mathrm{d}s},
\end{equation}
where $\sigma>0$ is the kernel bandwidth. Importantly, to carry out the statistical modeling, we must estimate $F_{i}^{-1}$ from the random sample $\{A_{ij}\}^{n_i}_{j=1}$. In practice, we use the empirical estimator $\widehat{F}_{i}(t)= \frac{1}{n_i} \sum_{j=1}^{n_i} 1\{A_{ij}\leq t \}$, $t\in \mathbb{R}$.

\subsection{Conditional Kernel Mean Embedding Estimation}
Our dataset is drawn from the NHANES 2011-2014 study, in which, for each individual \(i=1,\dots,n\), we record three demographic and anthropometric covariates
\[
  X_i = (\mathrm{sex}_i,\;\mathrm{weight}_i,\;\mathrm{age}_i)\in\mathbb{R}^3,
\]

\noindent
Here, \(\mathrm{sex}_i \in \{\mathrm{M}, \mathrm{F}\}\) denotes the biological sex, \(\mathrm{weight}\) is the body weight in kilograms, and \(\mathrm{age}_i\) is the age in years. For each subject \(i\), we consider a random response \(Y_i = \widehat{F}_i(\cdot) \in \mathcal{Y} = \mathcal{W}_2\), where \(\mathcal{W}_2\) denotes the 2-Wasserstein space of univariate probability distributions.

To model the relationship between covariates \(X\) and the distributional response \(Y,\) we employ kernel mean embeddings. Let
\[
  \psi:\mathcal X\to\mathcal H_X,\qquad
  \phi:\mathcal Y\to\mathcal H_Y
\]
\noindent be feature maps with associated kernels \(k_X\) and \(k_Y\), respectively. Given the training sample \(\{(X_i,Y_i)\}_{i=1}^n\) with all weights, we define the empirical risk
\[
  \mathcal R(C)
  = \frac{1}{n}\sum_{i=1}^n \bigl\|\phi(Y_i) - C\,\psi(X_i)\bigr\|_{\mathcal H_Y}^2
    + \lambda\,\|C\|_{\mathrm{HS}}^2,
\]
\noindent where \(C:\mathcal H_X\to\mathcal H_Y\) is a linear operator and \(\|\cdot\|_{\mathrm{HS}}\) denotes its Hilbert–Schmidt norm. By the representer theorem, the minimizer admits the closed‐form
\[
  \hat C
  = \Phi\,\bigl(K_X + n\lambda I_n\bigr)^{-1}\,\Psi^\top,
\]
with
\[
  \Phi = [\phi(Y_1),\dots,\phi(Y_n)],\quad
  \Psi = [\psi(X_1),\dots,\psi(X_n)],\quad
  (K_X)_{ij} = k_X(X_i,X_j).
\]
\noindent For a new covariate vector \(x\), the estimated conditional‐mean embedding is
\[
  \hat\mu_{Y\mid X}(x)
  = \hat C\,\psi(x)
  = \Phi\,(K_X + n\lambda I_n)^{-1}\,\mathbf{k}_x,
  \quad
  \mathbf{k}_x = \bigl(k_X(X_i,x)\bigr)_{i=1}^n.
\]

\subsection{Results}
\begin{figure}[h]
    \centering
  \includegraphics[width=0.9\textwidth]{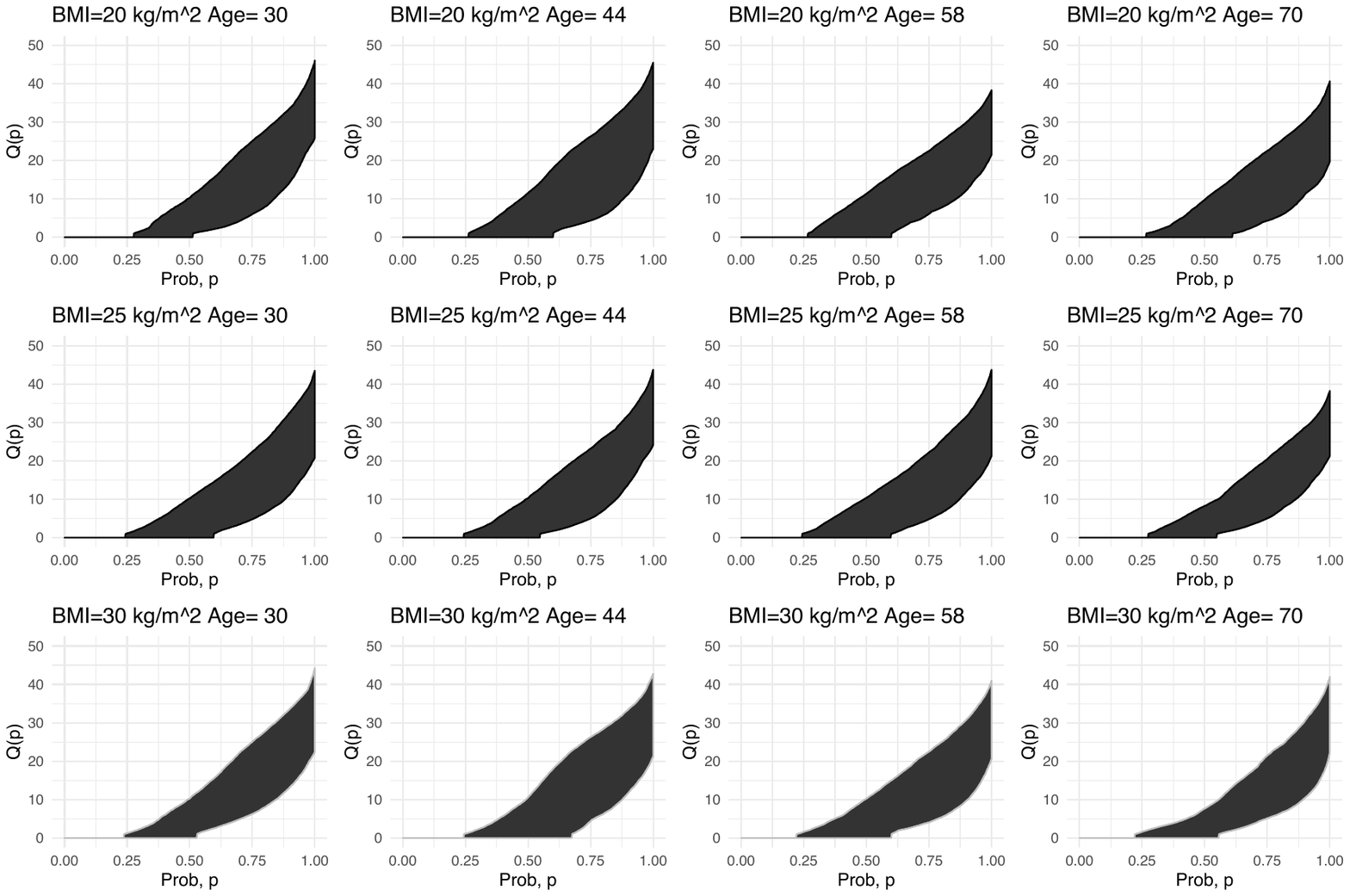}
    \caption{%
        (a) Women, BMI= $20\,kg/m^{2}$; %
        (b) Women, BMI= $25\,kg/m^{2}$; %
        (c) Women, BMI= $30\,kg/m^{2}$.%
        \newline
        Expected Quantile of Physical Activity Across Different Ages and Body Masses in Healthy Women of the American Population for $\alpha=0.5$.%
     \textcolor{red}{why not put more plots in each row? we are wasting a lot of space...} }
    \label{fig:tres_figuras}
\end{figure}

\noindent

\medskip

\medskip
\noindent\textbf{Recommended Physical Activity Values for Different Age and BMI Groups}

We apply the conformal algorithm to generate prediction regions for expected physical activity levels at a nominal confidence level of \(\alpha = 0.5\), adjusted for sex, age, and body mass index (BMI). For this analysis, we include only female participants without chronic diseases. Figure~\ref{fig:tres_figuras} displays the prediction regions for various combinations of age and BMI. Both the lower and upper bounds of the predicted activity values decline as age and BMI increase, underscoring the importance of tailoring physical activity recommendations to individual characteristics.

\medskip

\medskip
\noindent\textbf{Calibration Assessment}

Finally, to verify marginal coverage, we perform 10 random splits of the data into two equal halves. Across these splits, we obtain estimated coverage probabilities of approximately 0.92 for \(\alpha = 0.95\) and 0.51 for \(\alpha = 0.5\), demonstrating that our prediction regions are well calibrated and close to the nominal levels.

\section{Discussion}

This paper introduces a novel framework for uncertainty quantification in random statistical objects defined in separable Hilbert spaces, utilizing conformal prediction techniques and the depth measure derived from a kernel mean embedding. The efficacy of the new algorithms with finite--sample sizes has been thoroughly validated through an extensive simulation study, covering a spectrum of situations that include multivariate Euclidean data and functional responses. We also illustrate the algorithms in a physical activity application where functional information from accelerometer devices has been considered. The analysis of applications accentuates the necessity of creating specific individual decision-making to personalize physical guidelines according to patient characteristics.

Although the exposition within this manuscript predominantly focuses on the utilization of functional data, it is crucial to acknowledge the broader potential applicability of the methodology introduced to various complex data structures, including but not limited to graphs and trees, or any random object embeddings within a separable Hilbert space \cite{schoenberg1937certain,schoenberg1938metric}. The methods proposed here can be combined with any depth measure and not only with a kernel mean embedding (see for example \cite{hallin2021distribution}).

In future work, we focus on the extension of this kernel uncertainty quantification framework to the case of time-to-event data \cite{garcia2023causal} and dependent data.


\end{document}